\documentclass[letterpaper, 10 pt, conference]{ieeeconf}  

\IEEEoverridecommandlockouts                              

\usepackage{cite}
\usepackage{amsmath,amssymb,amsfonts}
\usepackage{algorithmic}
\usepackage{graphicx}
\usepackage{textcomp}
\usepackage{xcolor}
\usepackage{todonotes}
\usepackage{bm}
\usepackage{geometry}
\renewcommand{\vec}[1]{\bm{#1}}
\newcommand{\mat}[1]{\bold{#1}}

\title{\LARGE \bf
Lumped-Parameter Modeling and Control for Robotic High-Viscosity Fluid Dispensing in Additive Manufacturing
}

\author{William van den Bogert$^{1}$, James Lorenz$^{1}$, Xili Yi$^{2}$, Nima Fazeli$^{2}$, and Albert J. Shih$^{1}$
\thanks{$^{1}$ William van den Bogert, James Lorenz, and Albert J. Shih are with the Mechanical Engineering Department at the University of Michigan, MI, USA 
{\tt\small <willvdb,jplorenz,shiha>@umich.edu}}%
\thanks{$^{2}$ Xili Yi and Nima Fazeli are with the Robotics Department at the University of Michigan, MI, USA
{\tt\small <yixili,nfz>@umich.edu}}%
}

\begin{document}

\newgeometry{top=1in, bottom=0.75in, left=0.75in, right=0.75in}

\maketitle
\thispagestyle{empty}
\pagestyle{empty}

\begin{abstract}
In this paper, we present a novel flow model and compensation strategy for high-viscosity fluid deposition that yields high quality parts in the face of large transient delays and nonlinearity. Robotic high-viscosity fluid deposition is an essential process for a broad range of manufacturing applications including additive manufacturing, adhesive and sealant dispensing, and soft robotics. However, high-viscosity fluid deposition without compensation can lead to poor part quality and defects due to large transient delays and complex fluid dynamics. Our computationally efficient model is well-suited to real-time control and can be quickly calibrated and our compensation strategy leverages an iterative Linear-Quadratic Regulator to compute compensated deposition paths that can be deployed on most dispensing systems, without additional hardware. We demonstrate the improvements provided by our method when 3D printing using a robotic manipulator.

\end{abstract}

\section{Introduction}
The capability for precise robotic deposition of high-viscosity fluid is a key enabler for industries such as automotive manufacturing, tissue engineering, and soft robotics. When applied to 3D printing, this technology, known as direct ink writing (DIW) \cite{saa2022}, enables the production of high-quality flexible parts with complex geometries. The high viscosity of the fluid allows a printed shape to remain after deposition. However, this same viscosity leads to printing defects that require mitigation through compensation. The main contributions of this paper are a model that is quickly and easily generalizable for most pump-based high-viscosity fluid deposition systems, a calibration method for this model, and a demonstration of this model in a universally implementable open-loop control scheme without the need for additional hardware such as sensors.

\begin{figure}[t]
\centerline{\includegraphics[width=0.9\columnwidth]{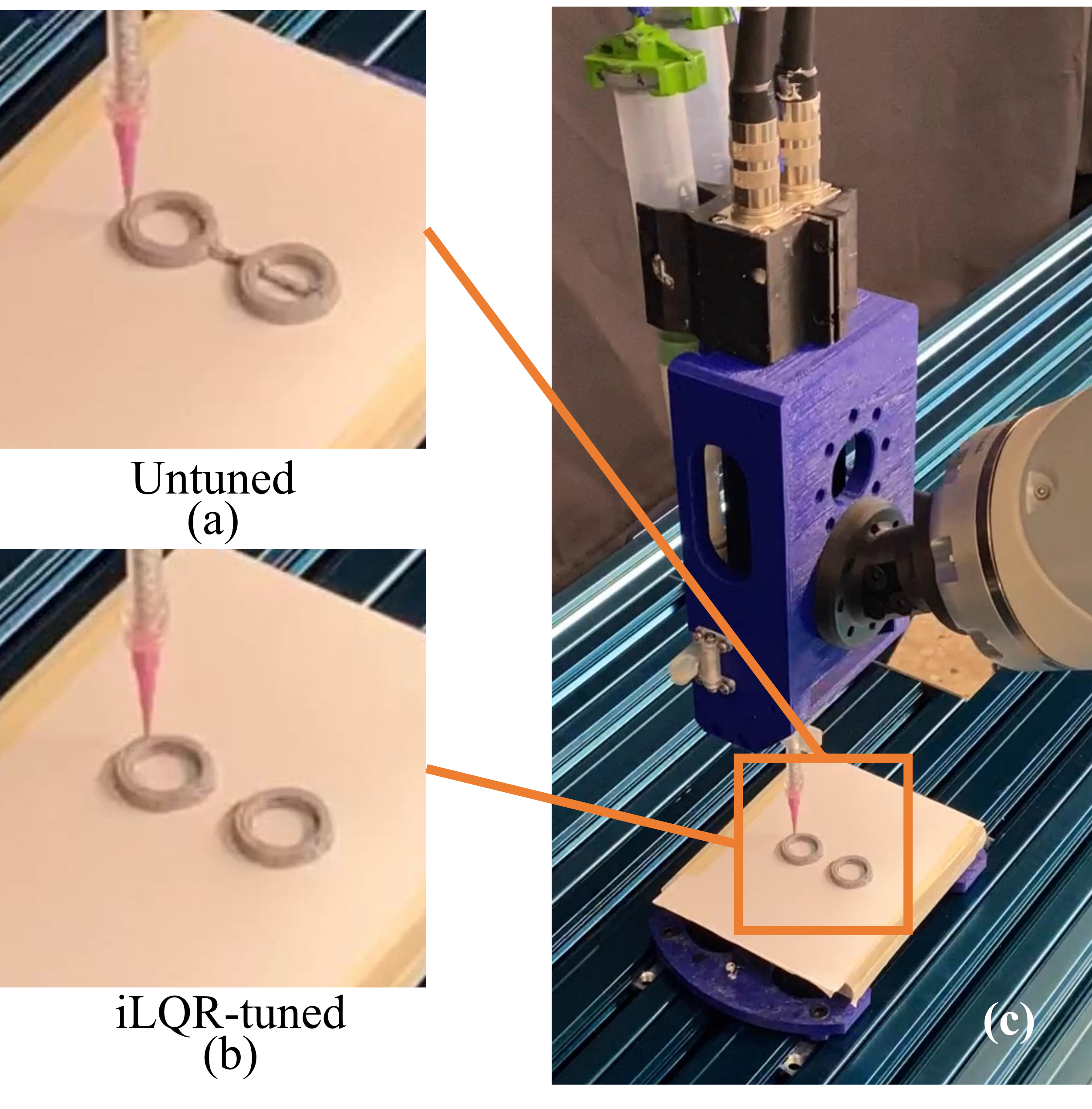}}
\caption{Robotic 3D printing with model-based flow compensation.
A silicone 3D printer head held by a robot arm printing two separated objects. Panel (a) shows the resultant print with untuned naive planning where we observe significant stringing and part defects due to over-extruded silicone. In comparison, Panel (b) 
shows the result of our approach with model-based flow compensation where the defect-free objects are separated cleanly. Panel (c) shows the robot 3D printing silicone.} \vspace{-5pt}
\label{fig:teaser}
\end{figure}

Our contributions enable DIW systems to approach the precision of more common 3D printing technologies. The traditional Fused Deposition Modeling 3D printing process typically controls flow behavior in a naive open-loop manner, where a motor position command controls the deployment of thermoplastic material. However, when the naive control is applied to DIW, transient effects due to viscosity-driven impedance produce imprecision in deposition that causes part defects such as the stringing found in Fig. \ref{fig:teaser}. 

Our work fills a gap in high-viscosity fluid deposition research with regards to describing the transient flow dynamics of high-viscosity fluids within the positive displacement pump (PDP) systems common to DIW. To model fluid flow behaviors of various PDP systems, research groups have developed control schemes using analytical solutions of physical laws \cite{Bar2016} \cite{Hap2021}, physics-based finite element simulations \cite{Jos2015} \cite{Fre2020} \cite{Pal2008}, and data-driven modeling approaches \cite{Fre2020} \cite{Zar2021}. However, these approaches are either too costly in numerical computation time for control applications, or do not consider transient effects in PDPs. To bridge this research gap, we propose a model of the fluid system dynamics that can predict transient effects of high-viscosity fluid flows in PDP systems, paired with an open-loop control paradigm to compensate for deposition errors during 3D printing. We demonstrate a significant improvement in fluid deposition precision when we apply this compensation to the trajectories for our robotic 3D printer, as shown in Fig. \ref{fig:teaser}.

\newgeometry{top=0.75in, bottom=0.75in, left=0.75in, right=0.75in}

\begin{figure}[]
\centerline{\includegraphics[width=\columnwidth]{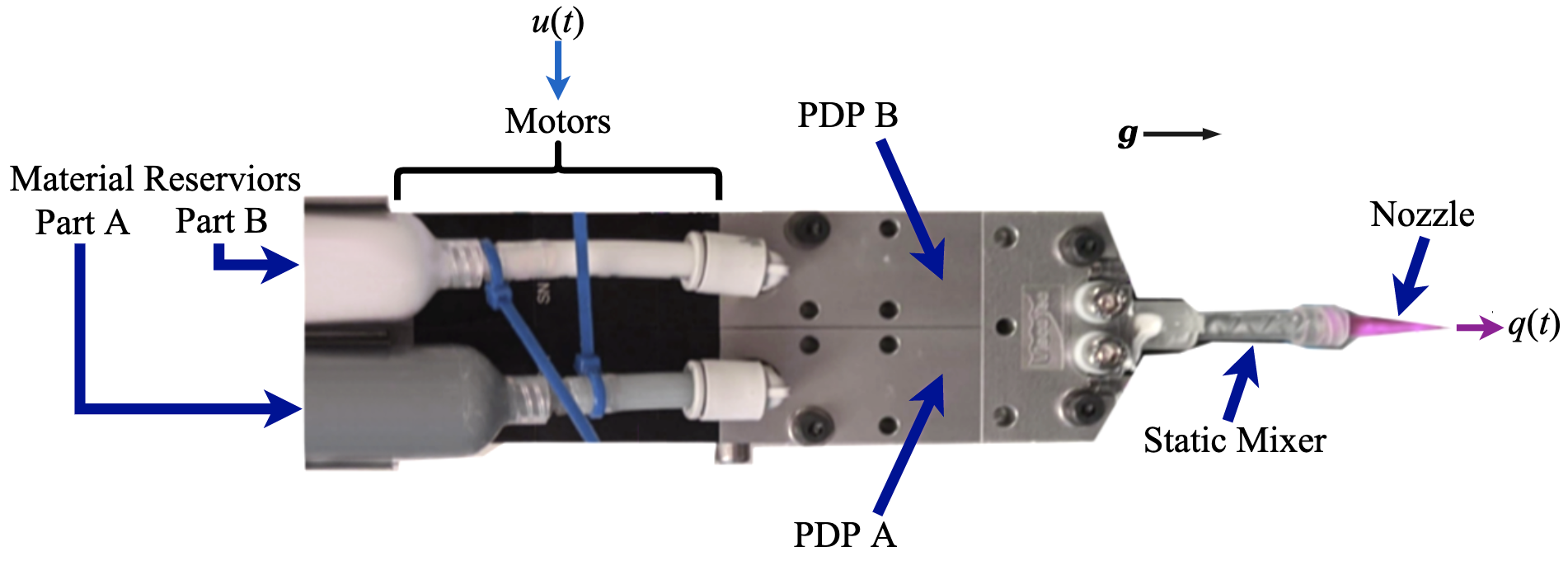}}
\caption{Positive displacement pump (PDP) system. Two progressive cavity pumps extrude A and B parts of the two-part liquid silicone. A and B parts are mixed via static mixing elements, and mixture is deposited through the nozzle. Here, $\vec{g}$ represents the direction of gravity.}
\label{fig：PDP}
\end{figure}

\subsection{Problem Statement}
The PDP system we use for fluid deposition is shown in Fig. \ref{fig：PDP}. We assume the fluids in reservoirs A and B share the same physical parameters, and that both motors receive the same input. Here, the control input is the motor speed signal $u(t)$. Motor speed is proportional to the intended rate of displacement of the progressing cavities in the PDPs. This input produces the nozzle outlet flow rate $q(t)$, and both $u(t)$ and $q(t)$ are in units of volumetric flow rate. While naive flow rate control assumes otherwise, we have observed that $u(t) \neq q(t)$ $\forall t$ (see Section \ref{section:results} for examples). For a given time horizon $n$, our goal is to solve for the trajectory $u(t)$ such that a prescribed $q(t)$ is achieved. We work with the discretized trajectories $u_k$ and $q_k$ over time steps $k=\lbrace1,...,n\rbrace$. Let $\mathcal{M}$ denote the model mapping the trajectories $\vec{u}_{n\times1}$ onto $\vec{q}_{n\times1}$:
\begin{equation}
\vec{q}_{model} = \mathcal{M}(\vec{u}) \label{eq:modelmap}
\end{equation}
This model $\mathcal{M}$ is an approximation of the true map $\mathcal{R}$ representing the complex nonlinear fluid dynamics. 
Our first contribution is developing a parameteric model $\mathcal{M}$ that is i) amenable to real-time control and ii) sufficiently closely approximates $\mathcal{R}$:
\begin{equation}
\mathcal{M}^* = \arg \min_{\mathcal{M}} |\mathcal{R}(\vec{u})-\mathcal{M}(\vec{u}) |\label{eq:modelprob}
\end{equation}
For open-loop control, we do not have access to $\mathcal{R}$, and instead use the model map $\mathcal{M}$ to find the optimal input trajectory $\vec{u}^*$ which produces the prescribed output trajectory $\vec{q}_{ref}$:
\begin{equation*}
\vec{u}^* = \arg \min_{\vec{u}} ||\mathcal{M}(\vec{u})-\vec{q}_{ref} ||\label{eq:ctrlprob}
\end{equation*}
subject to the constraints of the PDP pump and robot end-effector motion. The second contribution of this work is a strategy that leverages our proposed model to compute this optimal open-loop control.

\section{Related Works}

Baranovskii and Artemov have proposed generalized solutions for optimal control of nonlinear-viscous fluid models based on Navier-Stokes \cite{Bar2016}, though their specific analytic solutions demand simplifications specific to the fluid system. Hapanowicz's flow-resistance model \cite{Hap2021} is only applicable to quasi-steady cylindrical flows and does not account for transient pump dynamics. Froehlich and Kemmetmüller provide a computationally efficient control strategy \cite{Fro2020}, but it is designed only for valved systems found in injection molding machines.

To control the fluid flow of PDP systems, researchers instead developed simplified models with closed-loop controllers and numerical solving methods in place of analytical physics-based solutions of complex fluid models. Josifovic et al use computational fluid dynamics to model the quasi-step input of a valved pump system \cite{Jos2015}, and Wang et al combine pressure sensing equipment and PID optimization to correct errant flow behavior \cite{Wan2022}. Numerical solution methods have also been proposed for viscous flow control of screw-pump \cite{Xu2021}, helical gear pump \cite{Zha2018}, and peristaltic pump systems \cite{Lar2017}. Though seldom used in fluid dynamics controllers, a data-based lumped parameter model was used by Fresia and Rundo \cite{Fre2020} and Zardin et al \cite{Zar2021} to compensate pulsing fluid dynamics generated by variable displacement pumps and vane pumps, respectively. However, because each group derived their analytical flow models based on the geometry of their specific PDP system, their utility is limited because none are generalizable to the broad spectrum of PDP systems. 

Flow models for progressive cavity pumps are relatively unexplored in academic research, due to the significant improvement in accuracy that this system offers over alternative types of PDPs. Fisch et al \cite{Fis2020} and Franchin et al \cite{Fra2017} note a strong improvement in precision fluid deposition in 3D printing applications enabled by progressive cavity pump-controlled extrusion DIW systems. As a result, research on models for fluid interaction with progressive cavity PDPs has been limited to basic feedback control \cite{Woo2012} and numerical computational fluid dynamics methods to model specific behaviors of the progressive cavity pump system, including stator wear over time \cite{Pal2008}, inertial flow effects and viscous losses between the stator and rotor \cite{Pal2011}, and backflow of progressive cavity pumps \cite{Mul2021}. While these models predict unsteady flow in local regions of the progressive cavity systems (e.g. flow between the stators and rotors), none of these models provide a comprehensive control model of the entire fluid dispensing system required for our robotic 3D printing device, and all prior models require either additional hardware to close the control loop or computationally expensive solution methods which prohibit real-time compensation. 

\section{Methods}
Our approach consists of two main components: In Section \ref{subsection:modeling} we propose a novel computationally efficient high-viscosity fluid model that is amenable to real-time control and error compensation, and in Section \ref{subsection:control} we propose an optimal control formulation and solution method that leverages the model to rapidly compute open-loop control signals. 

\subsection{Modeling of High-Viscosity Fluid Deposition} \label{subsection:modeling}

\begin{figure}[]
\centerline{\includegraphics[width=\columnwidth]{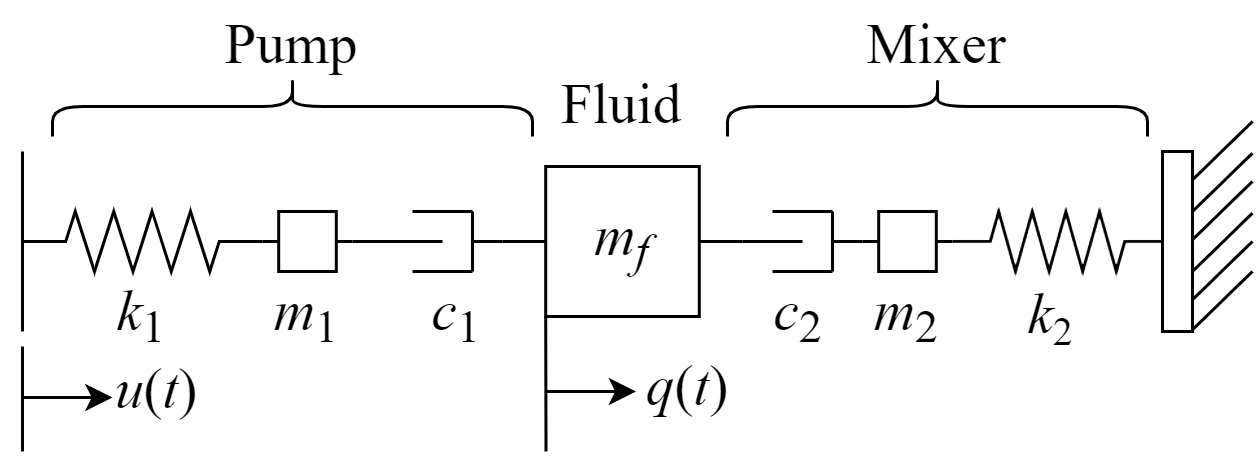}}
\caption{The high-viscosity fluid, pump and mixer are simplified to a 3 degree of freedom coupled linear dynamical system. Our model is a representation of fluid flow impedance in the pump (\(j=1\)) and mixer (\(j=2\)) in the form of elasticity \(k_j\), damping \(c_j\), and inertia \(m_j\), as well as inertia $m_f$ of the fluid.
}
\label{fig:lumped_parameter_structure}
\end{figure}

Our goal is to design $\mathcal{M}$ such that it i) is amenable to real-time control and ii) sufficiently closely approximates $\mathcal{R}$ in Eq. (\ref{eq:modelprob}), which represents the complex nonlinear fluid dynamics and interaction with both pump and mixer.
We model the entire PDP system as a 3 degree of freedom coupled linear dynamic system.
Our model is a representation of fluid flow impedance in the form of elasticity $k_j$, damping $c_j$, and inertia $m_j$ within both the pump ($j=1$) and mixer ($j=2$), as well as inertia $m_f$ of the fluid that sits between them. The model's mass-spring-damper system structure can be seen in Fig. \ref{fig:lumped_parameter_structure}.

When coupled with the initial state of the system, our map $\mathcal{M}$ introduced in Eq. (\ref{eq:modelmap}) can take the form of a dynamics function $f$ that predicts the next state vector $\vec{x}$, of which $q$ and its time derivative are elements:
\begin{equation}
\vec{x}_{k+1} = f(\vec{x}_k,u_k) \label{eq:dynfunc}
\end{equation}

The dynamics function defined in Eq. (\ref{eq:dynfunc}) is necessary for predictive modeling. The state vector $\vec{x}$ for the dynamics is defined using the 2 internal degrees of freedom $x_1$ and $x_2$ and the output degree of freedom $q$:
\begin{equation*}
  \vec{x} = \left[ {\begin{array}{cccccc}
    x_1 & x_2 & q & \dot{x_1} & \dot{x_2} & \dot{q} 
  \end{array} } \right]^T
\label{eq}
\end{equation*}
The system can be described with state-space form, with input $u$ and output $q$:
\begin{equation*}
\vec{x}_{k+1} = \mat{A}\vec{x}_{k}+\mat{B}u_k  \label{eq}
\end{equation*}
\begin{equation*}
q_k = \mat{C}\vec{x}_{k} \label{eq}
\end{equation*}
where the matrices $\mat{A}$, $\mat{B}$ and $\mat{C}$ are as follows:
\begin{equation*}
    \mat{A}_s=\left[ {\begin{array}{cccccc}
    0 & 0 & 0 & 1 & 0 & 0 \\
    0 & 0 & 0 & 0 & 1 & 0 \\
    0 & 0 & 0 & 0 & 0 & 1 \\
    -\frac{k_1}{m_1} & 0 & 0 & -\frac{c_1}{m_1} & 0 & \frac{c_1}{m_1} \\
    0 & -\frac{k_2}{m_2} & 0 & 0 & -\frac{c_2}{m_2} & \frac{c_2}{m_2} \\
    0 & 0 & 0 & \frac{c_1}{m_f} & \frac{c_2}{m_f} & -\frac{c_1+c_2}{m_f} \\\end{array} } \right] \label{eq}
\end{equation*}
\begin{equation*}
  \mat{A}=\mat{I}_{6\times6}+\mat{A}_s\Delta t \label{eq}
\end{equation*}
\begin{equation*}
  \mat{B}=\left[ {\begin{array}{cccccc}
    0 & 0 & 0 & \frac{k_1}{m_1} & 0 & 0 
  \end{array} } \right]^T \Delta t \label{eq}
\end{equation*}
\begin{equation*}
  \mat{C}=\left[ {\begin{array}{cccccc}
    0 & 0 & 1 & 0 & 0 & 0
  \end{array} } \right]^T \label{eq}
\end{equation*}

We identify values of the seven parameters shown in Fig. \ref{fig:lumped_parameter_structure} by using a gradient-descent scheme. Specifically, given measured data from the real system, we optimize the parameters so that the model response best-fits the data. The cost function $c$ for this optimization problem is calculated for a given time horizon as a function of the parameters $\vec{\phi}$, where the data is sampled with the same period as the simulation time-step $\Delta t$:
\begin{equation}
c(\vec{\phi}) = \sum_{k=1}^{n}\frac{1}{q_{k,meas}+q_{b}}(q_{k,model}(\vec{\phi})-q_{k,meas})^2 \label{eq:modelcost}
\end{equation}
\begin{equation*}
  \vec{\phi}=\left[ {\begin{array}{cccccccc}
    k_1 & c_1 & m_1 & m_f & k_2 & c_2 & m_2
  \end{array} } \right]^T \label{eq}
\end{equation*}
where $q_{k,meas}$ is the measured data for the time with index $k$, $q_{k,model}$ is the model flow rate for the time with index $k$, and $q_b$ is a biasing term. For each iteration of optimization $i$ the model is solved and the gradient of the cost computed. Then, the parameters are updated according to that gradient:
\begin{equation*}
\vec{\phi}_{i+1} = \vec{\phi}_{i}-\frac{h}{c(\vec{\phi}_i)}\frac{\partial{c(\vec{\phi_i})}}{\partial\vec{\phi}_i} \label{eq}
\end{equation*}
where $h$ is an adjustable increment. We set this optimization to stop once the cost changes by less than 0.1\% from one iteration to the next.

\subsection{Control of High-Viscosity Fluid Deposition}\label{subsection:control}

With this model in hand, we desire the input profile $u$ for a given time horizon such that the nozzle flow rate tracks the desired $q$. We construct a quadratic convex cost function $b$ for this optimization:

\begin{equation}
b_k = (\vec{x}_k-\vec{x}_k^{ref})^T\mat{Q}(\vec{x}_k-\vec{x}_k^{ref})+\vec{u}_k^T\mat{R}_k\vec{u}_k \label{eq:ilqrcost}
\end{equation}

where $\vec{x}_k$ is the current state, $u_k$ is the current input, $\vec{x}_k^{ref}$ is the desired state at the time with index $k$.

In our specific case, where our input only has a single degree of freedom, $\mat{R}_k$ and $\vec{u}_k$ reduce to scalars, so Eq. (\ref{eq:ilqrcost}) can be rewritten:
\begin{equation*}
b_k = (\vec{x}_k-\vec{x}_k^{ref})^T\mat{Q}(\vec{x}_k-\vec{x}_k^{ref})+R_ku_k^2 \label{eq}
\end{equation*}
where $\mat{Q}$ is a diagonal matrix containing the weightings for tracking error of the six states. Since we only are trying to track the output $q$, the third state, all other members of the diagonal can be set to zero:
\begin{equation*}
  \mat{Q}_{6\times6}=diag(0,0,\xi,0,0,0)
\end{equation*}

With the linear dynamics of our model, the cost function defined in Eq. \ref{eq:ilqrcost} could ordinarily be minimized analytically within the bounds of linear quadratic regulator (LQR) theory. However, we have observed that the reference trajectory is tracked more effectively in reality if the weighting on actuation effort $R_k$ is significantly reduced ($r_1<r_2$) once the input $u_k$ falls below a certain negative value $u_{th}$, which occurs when the pumps are run backwards in an effort to quickly reduce the output flow rate to zero when commanded as such:
\begin{equation*}
    R_{k} = \left\{
        \begin{array}{ll}
            r_1 & \quad u_k < u_{th} \\
            r_2 & \quad u_k \geq u_{th}
        \end{array}
    \right.
\end{equation*}

Since the impact of actuation effort on cost changes depending on the magnitude of the input, we must use an iterative LQR (iLQR) solver \cite{tod2005} for this optimization problem. Each iteration $i$ occurs in two parts. The first is a backward pass in which the controller gains $\mat{K}_k$ are computed using the value function $\mat{V}_k$ at the next time step, which is then updated for the current time step:
\begin{equation*}
\mat{K}_{k} = (R_k+\mat{B}^{*T}\mat{V}_{k+1}\mat{B}^*)^{-1}\mat{B}^{*T}\mat{V}_{k+1}\mat{A}^* \label{eq}
\end{equation*}
\begin{equation*}
\mat{P}_k = \mat{A}^*+\mat{B}^*\mat{K}_{k} \label{eq}
\end{equation*}
\begin{equation*}
\mat{V}_{k} = \mat{Q}^*_k+\mat{K}_{k}^TR_k\mat{K}_{k}+\mat{P}_k^T\mat{V}_{k+1}\mat{P}_k \label{eq}
\end{equation*}
where $\mat{A}^*$, $\mat{B}^*$, and $\mat{Q}^*_k$ are the homogenized versions of $\mat{A}$, $\mat{B}$, and $\mat{Q}$ respectively:
\begin{equation*}
\mat{A}^* = \left[ {\begin{array}{cc}
    \mat{A} & \mat{0}_{6\times1}\\
    \mat{0}_{1\times6} & 1\\
    \end{array} } \right], \quad \mat{B}^* = \left[ {\begin{array}{c}
    \mat{B}\\
    0\\
    \end{array} } \right] \label{eq}
\end{equation*}

\begin{equation*}
\mat{Q}^*_k = \left[ {\begin{array}{cc}
    \mat{Q} & \mat{Q}(\vec{x}_k-\vec{x}_k^{ref})\\
    (\vec{x}_k-\vec{x}_k^{ref})^T\mat{Q} & (\vec{x}_k-\vec{x}_k^{ref})^T(\vec{x}_k-\vec{x}_k^{ref}) \\
    \end{array} } \right] \label{eq}
\end{equation*}

The second part is a forward pass in which the input $u_k$ is updated according to the controller gains, and the state $\vec{x}_k$ is updated according to the dynamics with the input:
\begin{equation*}
u_{k}^{i+1}=u_k^{i}+\mat{K}_{k-1}(\vec{x}^{i+1}_{k-1}-\vec{x}^{i}_{k-1}) \label{eq}
\end{equation*}
\begin{equation*}
\vec{x}_{k+1}^{i+1}=\mat{A}\vec{x}_k^{i+1}+\mat{B}u_k^{i+1} \label{eq}
\end{equation*}

We set the controller optimization to stop if the cost changes by less than 0.1\% from one iteration to the next.

\section{Results}\label{section:results}

\subsection{Experimental Setup and Data Collection}

\begin{figure}[]
\centerline{\includegraphics[width=\columnwidth]{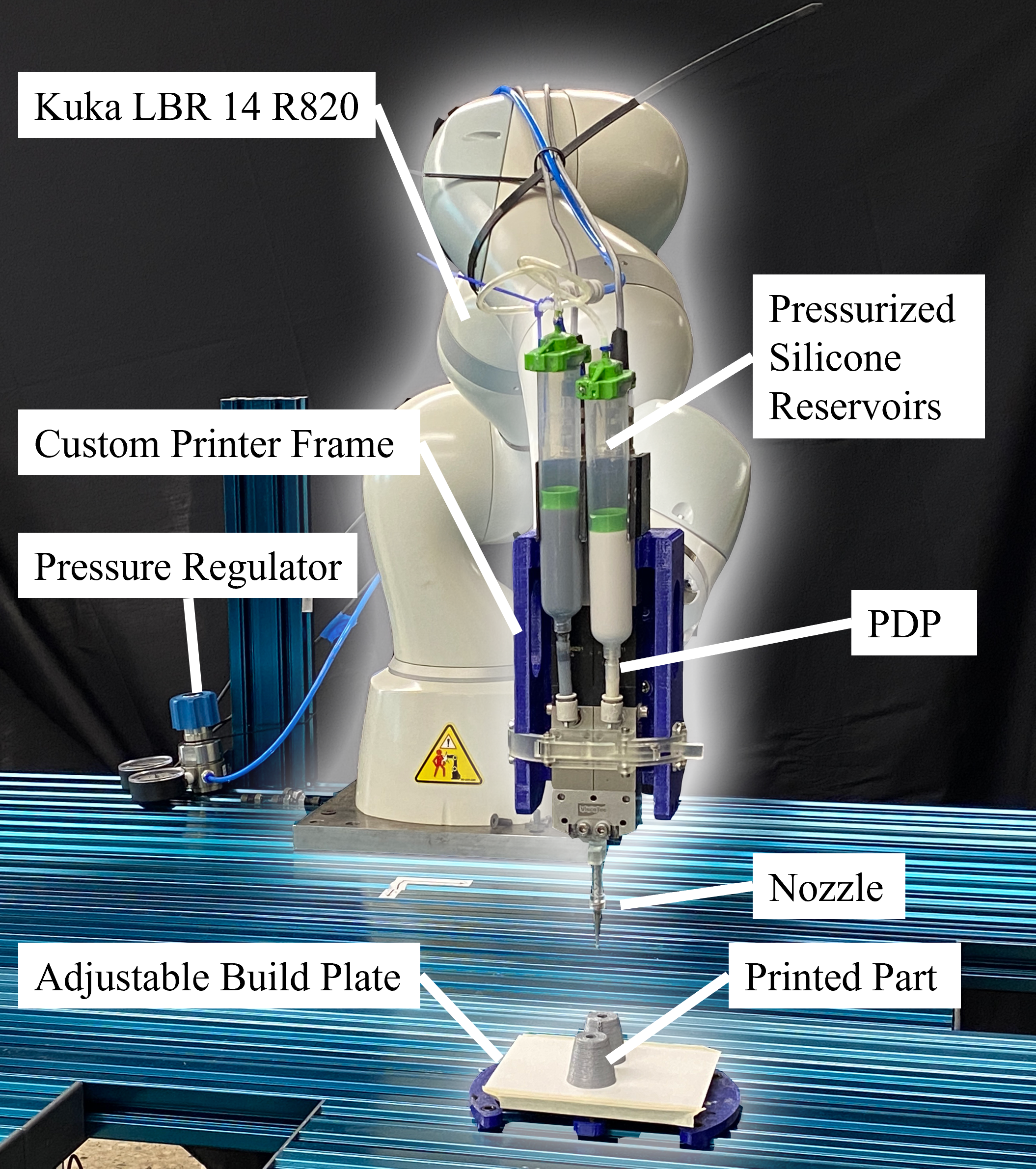}}
\caption{Experiment setup for printing. The whole robotic printing system consists of the robot arm, printer head, and adjustable build plate. The printer head includes a custom printer frame, pressurized silicone reservoirs, PDP, and a nozzle.}
\label{fig:experiment_setup}
\end{figure}

Our experimental setup can be seen in Fig. \ref{fig:experiment_setup}. We used a 7DOF Kuka LBR 14 R820 robot arm equipped with a customized print head alongside an adjustable build plate fixed to the base of the world frame. The customized print head consists of 3d-printed printer frame, pressurized reservoirs, PDPs, mixer, and nozzle. The silicone material used for 3D printing is Dow Corning 121 Structural Glazing Sealant. The stepper motors in the PDPs are controlled by a Raspberry Pi 4 Model B using integration with ROS, and the silicone reservoirs are connected to a high-pressure air source limited at 100 psi by a pressure regulator.

\begin{figure}
\centerline{\includegraphics[width=\columnwidth]{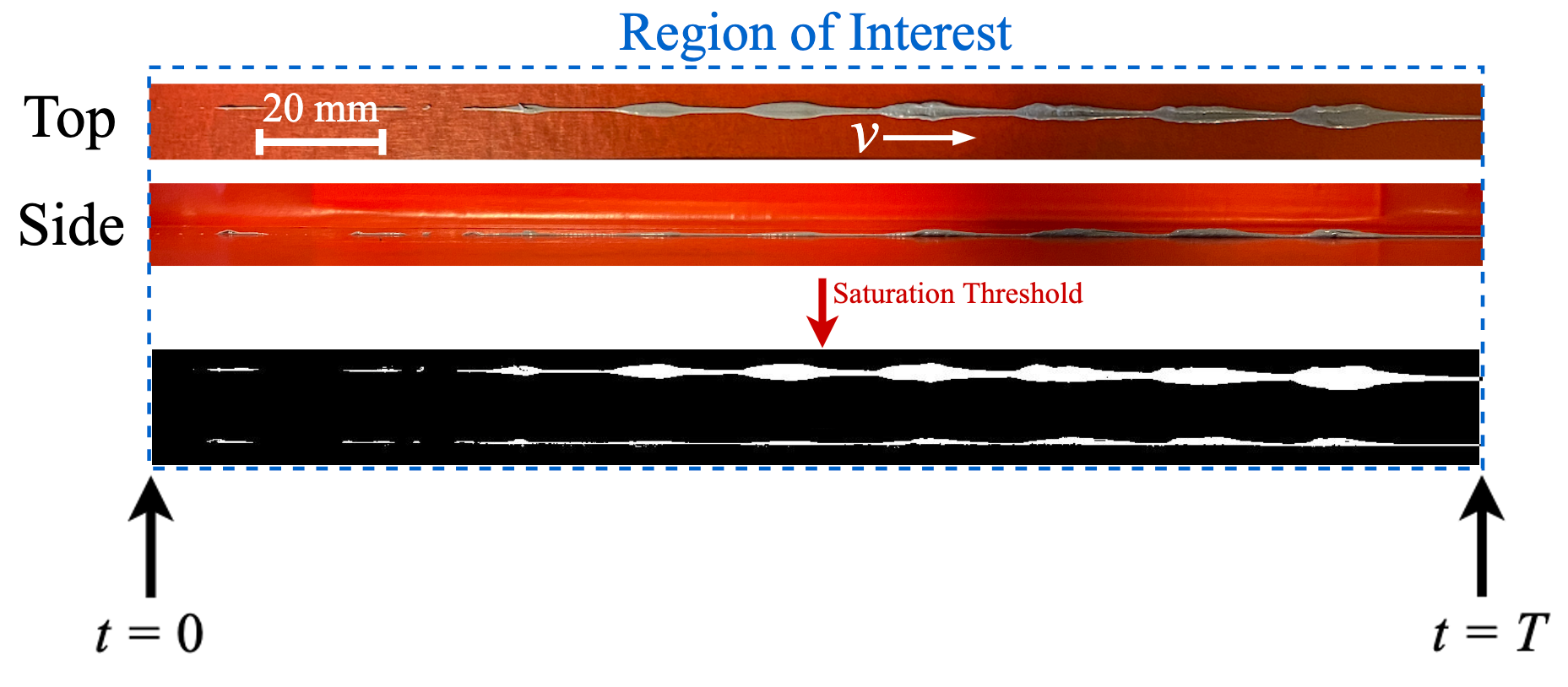}}
\caption{Visualization of binary segmentation measurement method for flow rate data. RGB images of both top and side views of printed beads are converted to HSV color space. We then segment them with a saturation threshold to get the flow rate measurement.}
\label{fig:measurement}
\end{figure}

We collected all data for model training and verification, as well as control verification, by segmenting side and top view images of a bead printed at constant velocity $v$ while varying commanded flow rate $u$, the result of which is shown in Fig. \ref{fig:measurement}. With the scale in Fig. \ref{fig:measurement}, we obtained a map between pixels and physical distance. Assuming that the cross-section of the printed bead is always rectangular, we calculated the cross-sectional area $A$ at each time step $k$. The nozzle flow rate is then: 
\begin{equation*}
q_{k,meas}=A_kv
\end{equation*}

\begin{figure}[h]
\centerline{\includegraphics[width=\columnwidth]{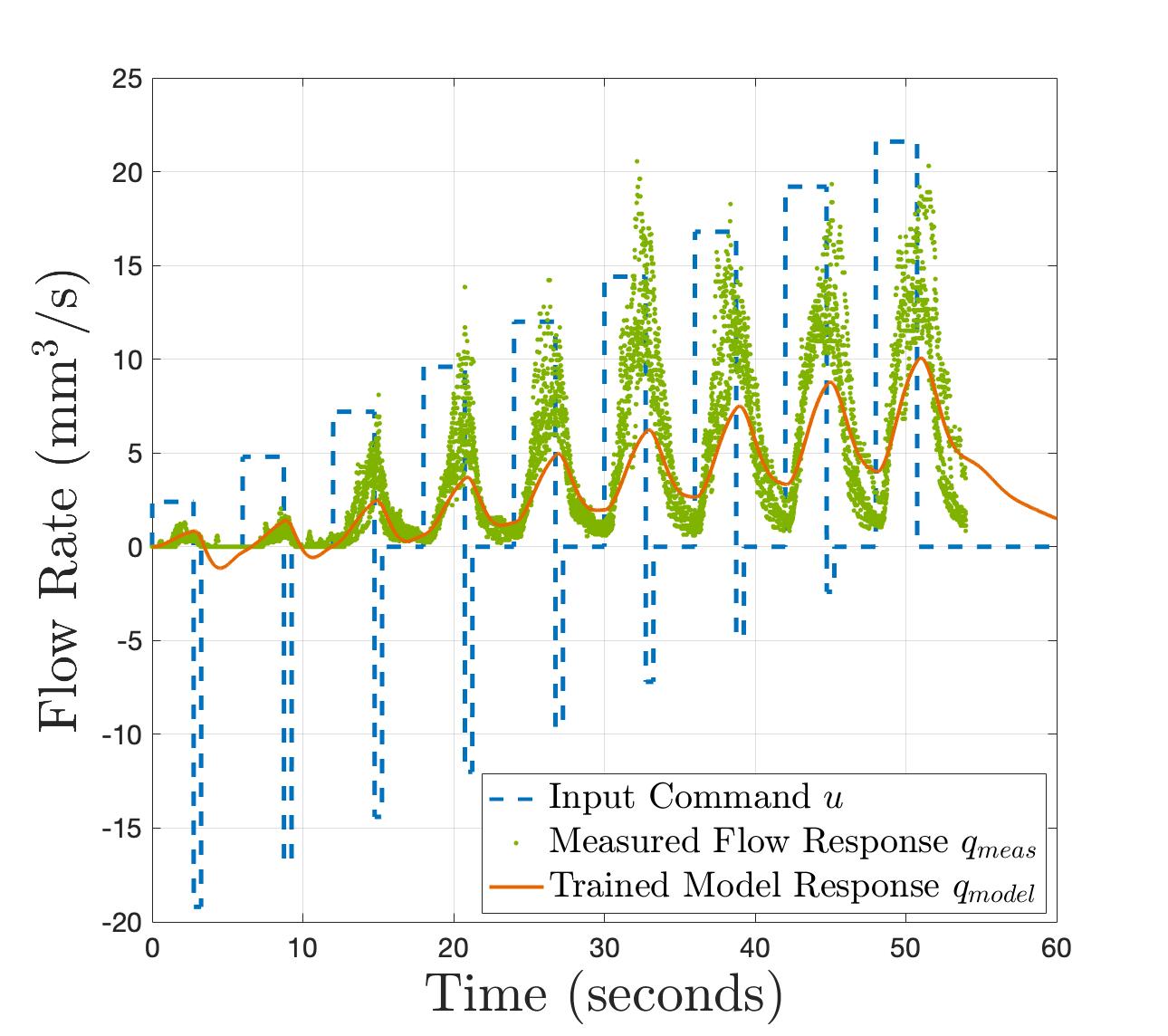}}
\caption{The results of training the lumped-parameter fluid deposition model show a more accurate prediction for lower magnitude flow rates. This is by design, using our selection of the $q_{b}$ term in Eq. (\ref{eq:modelcost}). Due to the real non-linear and specifically non-Newtonian nature of the silicone, a linear model cannot accurately predict across the entirety of a wide range of flow rate magnitudes.}
\label{fig:training_result}
\end{figure}

\subsection{Model Training}

To implement the model training algorithm described in Methods, we used values of $h=0.1$ for the step increment,  $q_{b} = 0.1$ mm$^3/$s for the cost function bias term,  $\Delta{t}=0.01$ s for the simulation time step. The data seen in Fig. \ref{fig:training_result} was sampled at this time step before the cost was computed.

We trained the model on an input command profile consisting of 18 square pulses of different magnitudes, both negative and positive, as seen in Fig. \ref{fig:training_result}. This was to anticipate the necessary distribution for compensating step-up and step-down flow rate profiles at the nozzle. The optimization returned the parameters shown in Table {\ref{tab1}}.

\begin{table}[h]
\begin{center}
\caption{Parameters found as a result of model training}
\begin{tabular}{l|l|l|l|l|l|l}
\hline
$k_1$ & $c_1$ & $m_1$ & $m_f$ & $k_2$ & $c_2$ & $m_2$\\ \hline
0.827 & 18.784 & 1.195 & 1.930 & 9.930 & 5.458 & 0.954 \\ \hline
\end{tabular}
\label{tab1}
\end{center}
\end{table}

With these parameters defined, the model exhibits the behavior shown in Fig. \ref{fig:training_result}. The trained model can be used to predict the data used in training, with the $q_{b}$ term chosen to ensure that lower magnitude flow rates are more accurately predictable. Higher magnitudes are thus less predictable. Fig. \ref{fig:prediction} shows the ability of the trained model to predict unseen data, consisting of the flow rate input and output during the targeted compensation of this paper.

\subsection{Controller Optimization}
To implement the iLQR solver described in the Methods section, we used values of $\xi=100$ for the tracking weight, $r_1=40$, $r_2=400$, $u_{th}=-2$ mm$^3/$s for the actuation effort weighting, and the time step was changed to $\Delta{t}=0.0005$ s.

\begin{figure}[h]
\centerline{\includegraphics[width=\columnwidth]{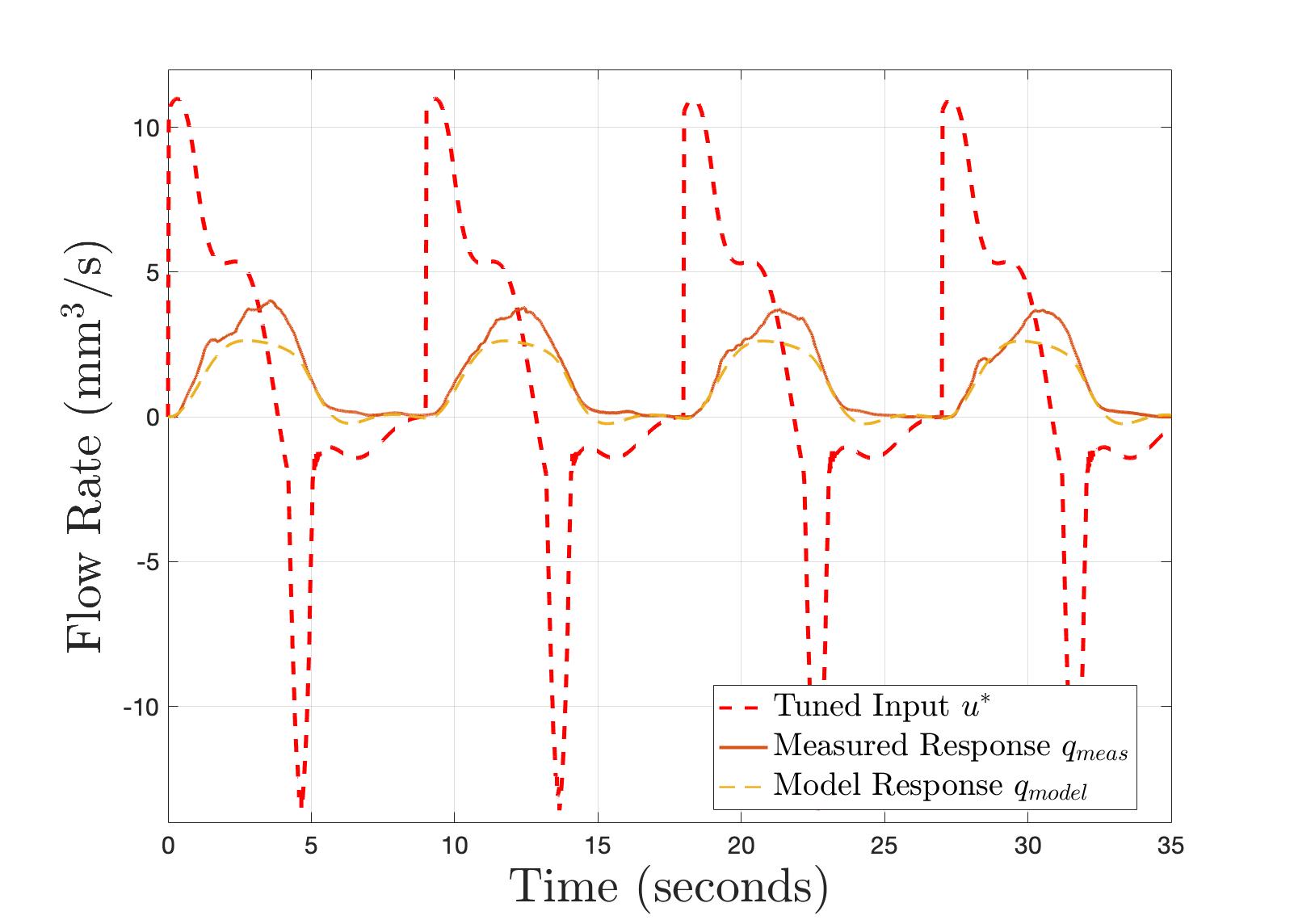}}
\caption{Prediction of unseen data by the lumped-parameter model. With the parameters from trained model, the model is able to predict unseen data, consisting of the flow rate input and output from the compensation validation.}
\label{fig:prediction}
\end{figure}

\begin{figure}[]
\centerline{\includegraphics[width=\columnwidth]{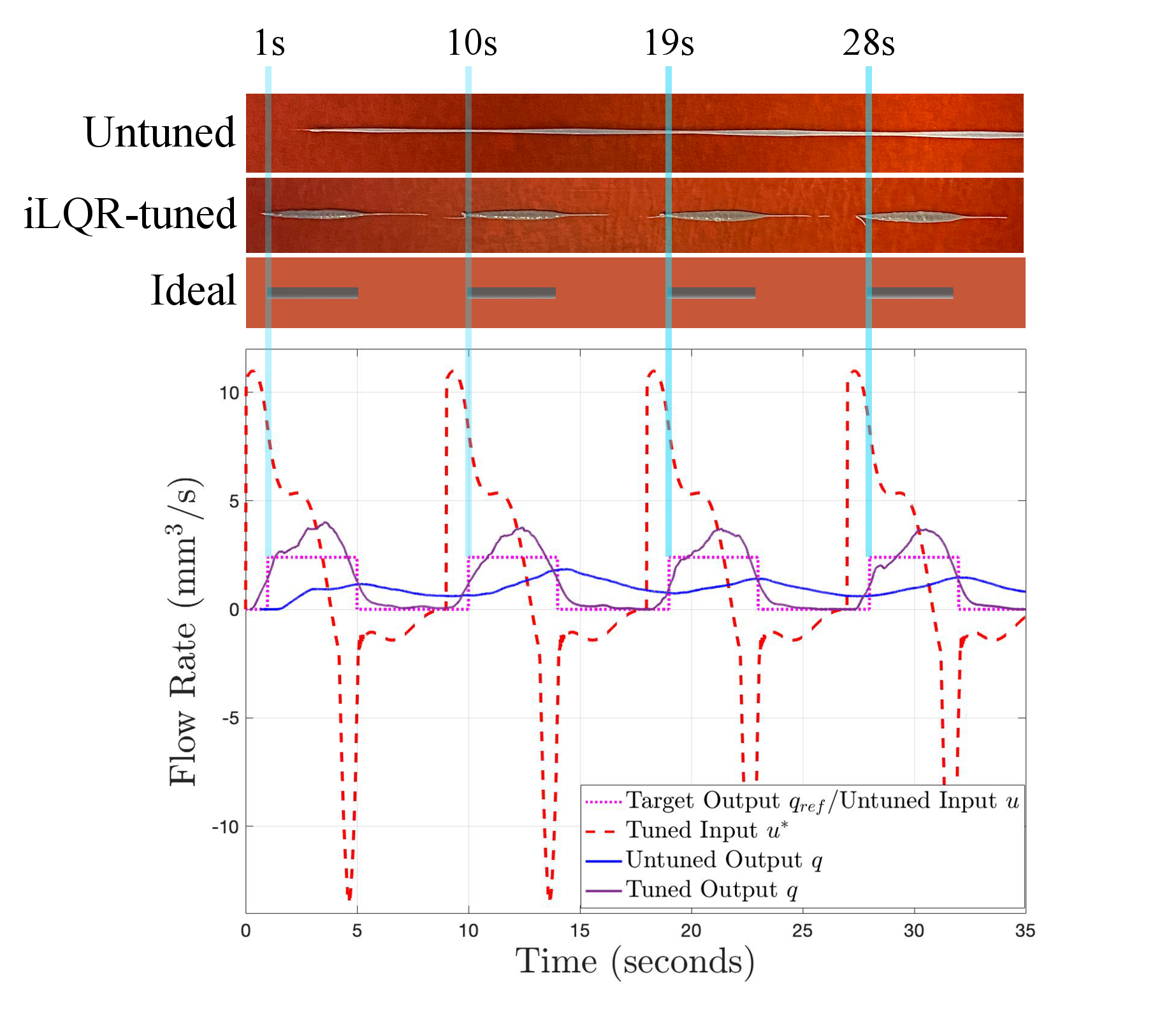}}
\caption{Comparison of model-based LQR control of fluid dispensing. Top pictures visualize the ideal, untuned, and iLQR-tuned fluid output profile. The bottom plot compares the input and output for them. The iLQR-tuned fluid output is closer to ideal output.}
\label{fig:comparison}
\end{figure}

Using these values, the iLQR solved for the input command seen in Fig. \ref{fig:prediction} and Fig. \ref{fig:comparison}. Fig. \ref{fig:comparison} shows the desired output flow profile, a series of four square pulses with a magnitude of 2.4 mm$^3$/s. These correspond to the four dashes shown in Fig. \ref{fig:comparison}, that would ideally be deposited as the nozzle travels from left to right. The naive open-loop control will involve these four square pulses being sent as an untuned input to the pumps. When we do this, however, the intention of depositing four discrete dashes is lost, due to the high impedance of deposition. When we use the iLQR-tuned input, the four dashes become much more distinct. The plot in Fig. \ref{fig:comparison} shows the flow rate vs. time response of the system provided with the untuned, and then iLQR-tuned input. The root-mean-square error between these respective responses and the desired response can be seen in Table \ref{tab2}, showing an improvement of over 50\% when the compensation is applied.

\begin{table}[]
\begin{center}
\caption{RMSE between desired output and measured output}
\begin{tabular}{l|l}
\hline
\textbf{Output}              & \textbf{RMSE} \\ \hline
\textit{Untuned response}    & 1.057 \\ \hline
\textit{iLQR-tuned response} & 0.522 \\ \hline
\end{tabular}
\label{tab2}
\end{center}
\end{table}

\subsection{Robotic Additive Manufacturing}

To validate the iLQR-tuned compensation during 3D printing, we devised printing tasks that would showcase misplaced material as a result of the parasitic effects when uncompensated. Trajectory for each printing task is generated with Matlab R2020b, and each waypoint in the trajectory includes the print head nozzle position in 3 cartesian coordinates and its corresponding flow rate. The print head is considered to only translate but not rotate. The robot and printing system communicate over a Robot Operating System (ROS) network where the robot sequentially executes waypoint while adjusting the flow rate with the Raspberry Pi connected to the network.

\begin{figure}[]
\centerline{\includegraphics[width=\columnwidth]{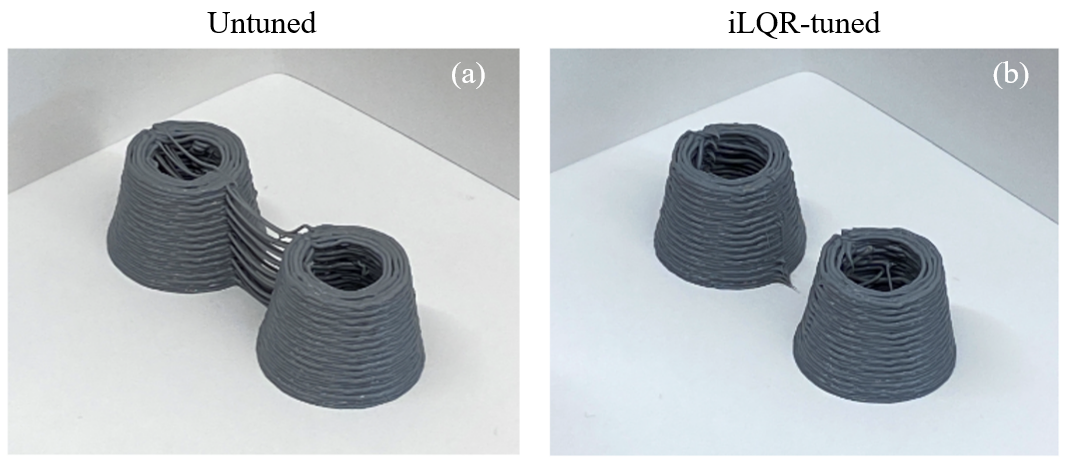}}
\caption{Robotic printing results. With untuned planning (a), over-extruded silicone was deposited between and inside the hollow truncated cones. With iLQR-tuned planning (b), the objects are separated cleanly.}
\label{fig:printing_results}
\end{figure}


We fabricated two printing samples with the same waypoints and target shape, that of two hollow truncated cones, separated by 40 mm between the central axes. One sample used the naive untuned flow rate, and the other used the iLQR-tuned flow rate. The printing results are shown in Fig. \ref{fig:printing_results}.

\begin{figure}[]
\centerline{\includegraphics[width=\columnwidth]{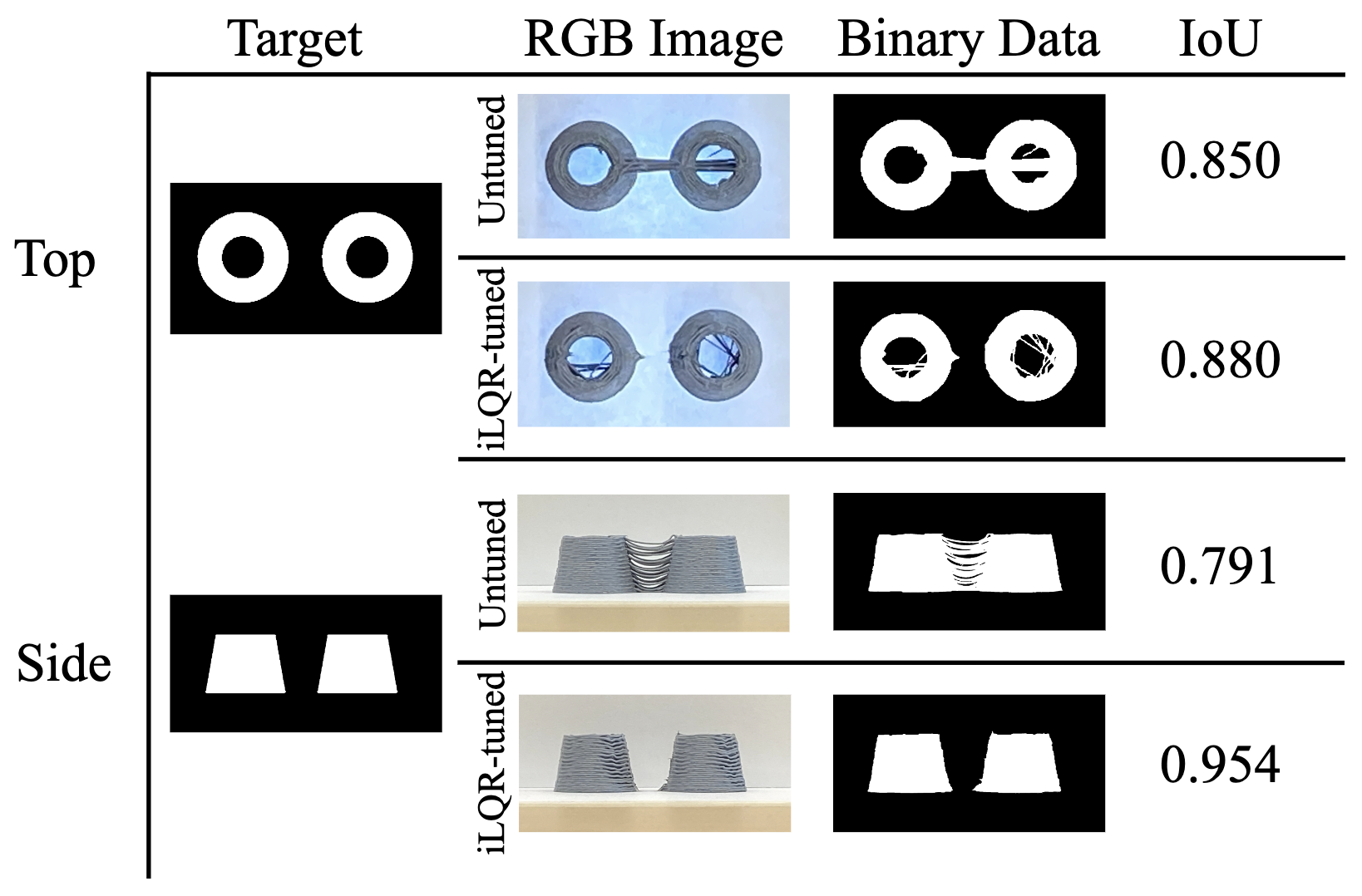}}
\caption{The segmentation and resulting intersection-over-union measurement, quantifying the success of the open-loop LQR control scheme over the naive control scheme.}
\label{fig:segementation_results}
\end{figure}

We segmented the top and side view RGB images of the printing results, producing binary masks of printed material, which can be seen in Fig.  \ref{fig:segementation_results}. We also produced a target binary mask (also in Fig. \ref{fig:segementation_results}) that represents the ideal placement of material, derived from the geometry of the mesh that we converted to waypoints for printing. To quantify the printing success, we calculated an intersection-over-union (IoU) score using the formula:
\begin{equation*}
    \text{IoU}(\mat{S}) = \frac{\mat{S}\cap{\mat{T}}}{\mat{S}\cup{\mat{T}}}
\end{equation*}
where $\mat{S} \in \lbrace0,1\rbrace_{y\times{z}}$ is the matrix representation of the binary mask in question (top or side view, iLQR-tuned or untuned) and $\mat{T} \in \lbrace0,1\rbrace_{y\times{z}}$ is the matrix representation of the associated target binary mask (top or side view).
The results shown in Fig. \ref{fig:segementation_results} demonstrate that iLQR-tuned printing has 0.030 and 0.163 higher IoU score than untuned printing when examining the top and side views respectively, demonstrating a more intentional placement of material.

\section{Discussion}

In this paper, we demonstrate how after quickly identifying just seven parameters that define our lumped-parameter fluid model, and applying the model to the iLQR flow control paradigm, we experimentally validated the efficacy of our approach with deposition tests using our robotic 3D printer. Misplaced material during the print was significantly reduced due to the application of our model-based compensation. Fig. \ref{fig:printing_results}(a) shows that with the untuned input, over-extruded silicone was deposited between the two hollow truncated cones, as well as inside the hollow region of one of the cones. In Fig. \ref{fig:printing_results}(b), the two objects are separated cleanly when the iLQR-tuned input is applied. Beyond additive manufacturing, the results in Fig. \ref{fig:comparison} show that the model-based compensation increases the precision of material placement along a single trajectory. This improvement has the potential to shrink the margin of error for robotic deposition of sealant and adhesive for manufacturing.

There are a few limitations with this work. Fig. \ref{fig:training_result} shows the model predicting less accurately for higher flow rate magnitudes. We selected the $q_{b}$ term so that the flow rates around the magnitude necessary for printing ($\sim$2–10 mm$^3/$s) would be better predicted by the model. The prediction of unseen data within this range can be seen in Fig. \ref{fig:prediction}, and it is satisfactory for the purposes of this paper. Due to our use of a linear model, the non-Newtonian behavior of the silicone cannot be predicted. Since iLQR can be easily used with nonlinear systems \cite{tod2005}, one approach for mitigating this limitation would be to use a more expressive model, such as a neural network, as a representation of the fluid dynamics. This may solve the eccentricities of the predictions seen in Fig. \ref{fig:prediction}.

One motivation to keep our fluid deposition model computationally inexpensive is to prepare for the future implementation of closed-loop control using such a model alongside vision-based feedback. Closed-loop control will provide further improvements in precise material placement, as our model cannot completely describe the real system. A more comprehensive control scheme will also take into account the dynamics of the robotic arm, optimizing the nozzle trajectory alongside the flow rate so that the transient delay still present in the tuned output (see Fig. \ref{fig:prediction}) will not be a permanent barrier to precise material placement. Through this we will achieve even further improvements than those demonstrated in this paper.


\bibliographystyle{ieeetr}
\bibliography{references.bib}

\end{document}